\documentclass{article}

\usepackage{microtype}
\usepackage{graphicx}
\usepackage{subcaption}
\usepackage{booktabs}
\usepackage{hyperref}

\usepackage[preprint]{icml2026}\usepackage{amsmath}
\usepackage{amssymb}
\usepackage{mathtools}
\usepackage{amsthm}
\usepackage{booktabs}
\usepackage{multirow}  
\usepackage{graphicx}
\usepackage{pifont} 

\usepackage{makecell}
\usepackage[utf8]{inputenc}
\usepackage{times}
\usepackage[most]{tcolorbox} 
\definecolor{grayBg}{HTML}{F2F2F2}
\newtcolorbox{positionbox}[1][]{
  enhanced,
  colframe=black!80,
  colback=grayBg,
  colbacktitle=blue!5!gray!10,
  coltitle=black,
  arc=5pt,
  boxrule=1.0pt,
  boxsep=0pt,  
  left=5pt,    
  right=5pt,   
  top=5pt,     
  bottom=5pt,  
  attach boxed title to top left={
    xshift=10pt,
    yshift=-\tcboxedtitleheight/2 
  },
  boxed title style={
    colframe=black!80,
    arc=3pt,
    boxrule=1.0pt,
  },
  #1
}
\newcommand{\cmark}{\ding{51}} 
\newcommand{\xmark}{\ding{55}} 
\usepackage{colortbl} 
\usepackage[capitalize,noabbrev]{cleveref}

\theoremstyle{plain}

\theoremstyle{definition}

\theoremstyle{remark}

\makeatletter
\renewcommand{\icmlcorrespondingauthor}[2]{%
  \ifdefined\icmlcorrespondingauthor@text
    \g@addto@macro\icmlcorrespondingauthor@text{, #1#2}%
  \else
    \gdef\icmlcorrespondingauthor@text{#1#2}%
  \fi
}

\renewcommand{\printAffiliationsAndNotice}[1]{\global\icml@noticeprintedtrue%
  \stepcounter{@affiliationcounter}%
  {\let\thefootnote\relax\footnotetext{\hspace*{-\footnotesep}\ificmlshowauthors #1\fi%
      \forloop{@affilnum}{1}{\value{@affilnum} < \value{@affiliationcounter}}{
        \textsuperscript{\arabic{@affilnum}}\ifcsname @affilname\the@affilnum\endcsname%
          \csname @affilname\the@affilnum\endcsname%
        \else
          {\bf AUTHORERR: Missing \textbackslash{}icmlaffiliation.}
        \fi
      }.%
      \ifdefined\icmlcorrespondingauthor@text
          Contact: \icmlcorrespondingauthor@text.
      \else
        {\bf AUTHORERR: Missing \textbackslash{}icmlcorrespondingauthor.}
      \fi

      \ \\ 
      \Notice@String
    }
  }
}
\makeatother

\usepackage[textsize=tiny]{todonotes}

\icmltitlerunning{
	Position: Beyond Model-Centric Prediction—Agentic Time Series Forecasting
}

\begin{document}
	
	\twocolumn[
	\icmltitle{
		Position: Beyond Model-Centric Prediction—Agentic Time Series Forecasting
	}
	
	
	
	\icmlsetsymbol{equal}{*}
	
	\begin{icmlauthorlist}
		\icmlauthor{Mingyue Cheng}{yyy}
		\icmlauthor{Xiaoyu Tao}{yyy}
		\icmlauthor{Qi Liu}{yyy}
		\icmlauthor{Ze Guo}{yyy}
		\icmlauthor{Enhong Chen}{yyy}
	\end{icmlauthorlist}
	
	\icmlaffiliation{yyy}{State Key Laboratory of Cognitive Intelligence, University of Science and Technology of China Hefei, Anhui Province, China}
	
	\icmlcorrespondingauthor{}{qiliuql@ustc.edu.cn}
	\icmlcorrespondingauthor{}{mycheng@ustc.edu.cn}
	\icmlkeywords{Machine Learning, ICML}
	
	\vskip 0.3in
	]
	
	\printAffiliationsAndNotice{}
	
	\begin{abstract}
		Time series forecasting has traditionally been formulated as a model-centric, static, and single-pass prediction problem that maps historical observations to future values. While this paradigm has driven substantial progress, it proves insufficient in adaptive and multi-turn settings where forecasting requires informative feature extraction, reasoning-driven inference, iterative refinement, and continual adaptation over time. In this paper, we argue for agentic time series forecasting (ATSF), which reframes forecasting as an agentic process composed of perception, planning, action, reflection, and memory. Rather than focusing solely on predictive models, ATSF emphasizes organizing forecasting as an agentic workflow that can interact with tools, incorporate feedback from outcomes, and evolve through experience accumulation. We outline three representative implementation paradigms—workflow-based design, agentic reinforcement learning, and a hybrid agentic workflow paradigm—and discuss the opportunities and challenges that arise when shifting from model-centric prediction to agentic forecasting. Together, this position aims to establish agentic forecasting as a foundation for future research at the intersection of time series forecasting. \footnote{\url{https://github.com/Mingyue-Cheng/atsf}}
	\end{abstract}
	
	\section{Introduction}
Time series forecasting plays a central role in decision-making across a wide range of real-world applications~\cite{box2015time,januschowski2020criteria}, including energy system operation, healthcare management, financial risk control. In these settings, forecasts are rarely consumed as isolated numerical outputs; instead, they serve as critical evidence that informs downstream decisions such as resource allocation, policy adjustment, and risk mitigation. The value of forecasting therefore lies not only in predictive accuracy, but also in how effectively forecasts support decision processes under uncertainty, evolving conditions, and operational constraints~\cite{zhang2024probts}. Understanding forecasting as an core component of decision-making provides an essential perspective for re-examining how forecasting systems should be formulated and evaluated in practice~\cite{zhang2025decision}.
	
Most existing research on time series forecasting has been developed under a model-centric paradigm~\cite{torres2021deep,cheng2023timemae,hollmann2025accurate,chow2024towards}, where forecasting is typically formulated as a supervised learning problem that maps historical observations to future values. Under this formulation, a forecasting task is defined by a fixed input, a predetermined prediction horizon, and a single-pass execution of a predictive model. Progress in this area has largely focused on designing more expressive model architectures~\cite{wen2022transformers,salinas2020deepar,lim2021temporal,wu2021autoformer,nie2022time,rasul2021autoregressive}, improving representation learning, and scaling training procedures, leading to substantial empirical advances across a wide range of benchmark datasets~\cite{aksu2024gift,li2025tsfm}. Within this paradigm, the forecasting process is assumed to be fully specified once the data, model, and objective are defined, and forecasting performance is primarily attributed to the predictive capacity of the designed model~\cite{brigato2025position}.
	
However, real-world forecasting practice often exhibits a structural mismatch with this single-pass formulation~\cite{garza2025timecopilot}. In many practical settings, forecasting is inherently adaptive and multi-turn~\cite{zhang2025alphacast}, involving a sequence of connected steps rather than a one-time model execution. Before generating predictions, practitioners usually interpret raw data, select relevant features, and determine which contextual information is most informative for the task at hand~\cite{roque2025cherry}. After forecasts are produced, results are typically evaluated, questioned, and revised in light of observed discrepancies, uncertainty, or domain knowledge~\cite{liu2025improving,xie2025causality}. Over repeated use, forecasting systems are further expected to adapt to changing environments, transfer experience across instances, and refine their forecasting strategies over time. These aspects—informative feature extraction~\cite{cheng2026instructtime++}, reasoning-driven inference~\cite{cheng2025can,kong2025position}, iterative refinement~\cite{zhang2025alphacast}, and continual evolving—are central to effective forecasting in practice, yet they are only implicitly or partially captured by conventional model-centric formulations.
	
\vspace{-0.1in}
\paragraph{Our Position.}To better align forecasting formulations with the characteristics of adaptive and multi-turn practice, we propose agentic time series forecasting (ATSF) as a conceptual reframing of the forecasting problem. ATSF views forecasting not as a single-pass predictive computation, but as an agentic process composed of interrelated decision steps, including perception, planning, action, reflection, and memory. Under this perspective, forecasting is organized as an multi-turn interaction process in which objectives are formulated, contextual information is prepared, predictions are generated through interaction with tools, outcomes are evaluated and revised, and experience is accumulated over time. Rather than attributing forecasting performance solely to the capacity of a predictive model, ATSF emphasizes how forecasting activities are structured, coordinated, and adapted through this agentic modeling process. By shifting the focus from model execution to process organization, ATSF provides a principled framework for understanding forecasting as a dynamic, decision-driven activity that evolves through interaction and experience.

\vspace{-0.1in}
\paragraph{Contributions.}This paper presents a position on ATSF with the goal of clarifying its conceptual foundations and outlining a forward-looking research agenda. Rather than introducing a specific forecasting model, we formalize ATSF as a paradigm that reframes forecasting as an reasoning-driven and iterative refinement process. We examine why such a reframing is necessary, define the core components of agentic forecasting, and discuss multiple implementation paradigms that instantiate this perspective in practice. Finally, we identify key opportunities and challenges that arise when moving from model-centric prediction to agentic forecasting. Through this discussion, we aim to provide a unified conceptual framework that can guide future research at the intersection of time series forecasting.
	

\begin{table*}[t]
	\centering
	\caption{Evolution of time series forecasting paradigms: from statistical models to agentic llms.}
    \small
	\label{tab:ts_paradigm_evolution}
	\resizebox{\textwidth}{!}{%
    \renewcommand{\arraystretch}{1}
		\begin{tabular}{c c >{\centering\arraybackslash}m{5.3cm} c c c c c c}
			\toprule
			\textbf{Paradigm} & \textbf{Modeling Idea} & \textbf{Rep. Works} & \textbf{General.} & \textbf{Effic.} & \textbf{Train} & \textbf{Tool} & \textbf{Evol.} & \textbf{Interp.} \\
			\midrule
			Statistical Modeling & \makecell{Handcrafting parametric assumptions \\ to model temporal dynamics} & ARIMA~\cite{hyndman2008automatic}, CES~\cite{svetunkov2022complex}, Prophet~\cite{taylor2018forecasting} & Low & High & \cmark & \xmark & \xmark & High \\
            \midrule
			Machine Learning & \makecell{Manually extracting extensive \\ features from raw data} & XGBoost~\cite{chen2016xgboost}, LightGBM~\cite{ke2017lightgbm}, SVR~\cite{terrault2016management} & Low & High & \cmark & \xmark & \xmark & Med \\
            \midrule
			Deep Learning & \makecell{Leveraging expressive network architecture\\ to learning powerful representations} & TFT~\cite{lim2021temporal}, DeepAR~\cite{salinas2020deepar}, Informer~\cite{zhou2021informer}, PatchTST~\cite{Yuqietal-2023-PatchTST}, ConvTimenet~\cite{cheng2025convtimenet}, MTGNN~\cite{wu2020connecting} & Med & Med & \cmark & \xmark & \xmark & Low \\
            \midrule
			Foundation Models & \makecell{Learning universal representations \\ through large-scale pretraining \\ across diverse tasks and domains} & Chronos~\cite{ansari2024chronos}, TimesFM~\cite{das2024decoder}, Sundial~\cite{liu2025sundial} & High & Low & \cmark & \xmark & \xmark & Low \\
            \midrule
			\makecell{LLM-based \\ Generative Models} & \makecell{Align pre-trained LLMs to \\ time series via  SFT or RL} & Time-LLM~\cite{jintime}, TokenCast~\cite{tao2025values}, LLMTime~\cite{gruver2023large}, PromptCast~\cite{xue2023promptcast} & High & Low & \cmark & \xmark & \xmark & Med \\
            \midrule
            \makecell{Reasoning-Driven \\ Forecasting  Models} & \makecell{Formulating forecasting as \\ an explicit reasoning process \\ over historical observations and context} & TS-Reasoning~\cite{chow2024towards}, TimeOmni-1~\cite{guan2025timeomni}, Time-R1~\cite{luo2025time}, TimeReasoner~\cite{cheng2025can}  & High & Low & \xmark & \xmark & \xmark & Med \\
			\midrule
		 \textbf{ATSF} & \makecell{\textbf{Reasoning-driven action} \\ \textbf{and iterative refinement } \\ \textbf{via autonomous multi-turn interaction}} & TimeCopilot~\cite{garza2025timecopilot}, Timeseriesscientist~\cite{zhao2025timeseriesscientist}, CastMind~\cite{zhang2025alphacast}, Cast-R1~\cite{tao2026cast}  & \textbf{High} & \textbf{Low} & \xmark & \cmark & \cmark & \textbf{High} \\
			\bottomrule
		\end{tabular}%
	}
\end{table*}

\section{Why Agentic Time Series Forecasting}
This section explains why moving beyond model-centric prediction toward agentic forecasting is a principled and necessary step, driven by both the intrinsic nature of forecasting tasks and recent advances in intelligent systems.

\subsection{The Limitations of Model-Centric Prediction}
Despite substantial gains in predictive accuracy, model-centric forecasting approaches suffer from fundamental limitations that arise from problem formulation rather than insufficient model capacity. As summarized in Table~\ref{tab:ts_paradigm_evolution}, dominant paradigms in time series forecasting—from statistical modeling and machine learning to deep learning, foundation models, and recent LLM-based adaptations—employ different modeling techniques but consistently organize forecasting around model-centric prediction, where progress is driven by improved representations, scale, or training procedures. Across these paradigms, forecasting is formulated as a single-pass execution with fixed inputs and predefined objectives. Although later approaches improve the quality of representation, none explicitly incorporate planning, tool interaction, iterative self-correction, or experience accumulation as core components of the forecasting process. These formulation-level characteristics lead to structural limitations that become particularly evident in adaptive, long-term, and decision-oriented settings.

These limitations manifest along several dimensions that directly shape how forecasting systems are designed and used in practice. First, model-centric forecasting is inherently static. Once inputs, features, and prediction horizons are specified, the task proceeds without mechanisms for revisiting objectives or restructuring the problem, limiting the ability to adjust goals, decompose complex tasks, or respond to dynamic contexts. Second, model-centric prediction operates in a largely closed manner, with limited capacity for autonomous interaction with external tools. Forecasting is typically confined to a single predictive model, restricting the integration of complementary analyses, domain knowledge, or external information in complex scenarios. Third, conventional forecasting is predominantly single-pass, lacking intrinsic mechanisms for iterative evaluation and revision. Predictions are generated once and treated as final outputs, with no internal support for reflection, assumption checking, or corrective reasoning, constraining iterative improvement and effective decision support. Finally, model-centric approaches do not explicitly support experience accumulation over time. Knowledge from past forecasting instances is absorbed indirectly through retraining or parameter updates rather than being represented and reused, limiting experience transfer, long-term adaptation, and continual evolution in changing environments.

\subsection{Why Forecasting Must Be Agentic}
The necessity of ATSF arises from both the intrinsic properties of forecasting tasks and recent advances in intelligent systems capable of supporting agentic reasoning and interaction~\cite{zhang2024self,sibai2026path,xu2026ai}. Together, these factors suggest that reframing forecasting as an agentic process is not merely a design choice, but a principled response to the nature of the problem and the capabilities of modern artificial intelligence (AI).

First, the properties of time series data and the nature of forecasting tasks inherently demand agentic reasoning~\cite{wei2026agentic}. Time series data are often noisy, non-stationary, and context-dependent, with patterns evolving across time and domains. Effective forecasting therefore requires more than fitting a function to historical observations; it involves deciding which information is relevant, how objectives should be formulated, and when predictions should be questioned or revised. In practice, forecasting is rarely a one-shot inference problem. Instead, it unfolds as a process involving interpretation, hypothesis formation, iterative adjustment, and experience accumulation. These characteristics resemble how human experts forecast—by perceiving new information, planning strategies, acting through informed decisions, reflecting on outcomes, and updating their understanding over time. Capturing this process-level complexity necessitates an agentic formulation in which forecasting decisions are explicitly represented and iteratively refined.

Second, recent advances in large language model (LLM) make such a formulation technically viable. Modern LLM-based agents exhibit emerging capabilities in reasoning, planning, tool use, and self-reflection, enabling behavior beyond single-pass generation. Systems such as conversational agents and deep research frameworks~\cite{xu2025comprehensive,cheng2026mind2report} demonstrate how complex tasks can be decomposed, executed through interaction, and refined over multiple steps with accumulated memory. These developments open a design space in which forecasting can be organized as an interactive, multi-step decision process rather than a static single-pass model execution. Importantly, the relevance of agentic forecasting does not depend on any specific model architecture, but on the availability of systems that can coordinate perception, planning, action, reflection, and memory within a unified agentic workflow.

\section{What is Agentic Time Series Forecasting}
We will formalize ATSF by defining its core concepts, key components, and the iterative decision-making process that distinguishes it from model-centric forecasting paradigms.	

\begin{figure*}[t]  
    \centering
    \includegraphics[width=1.95\columnwidth]{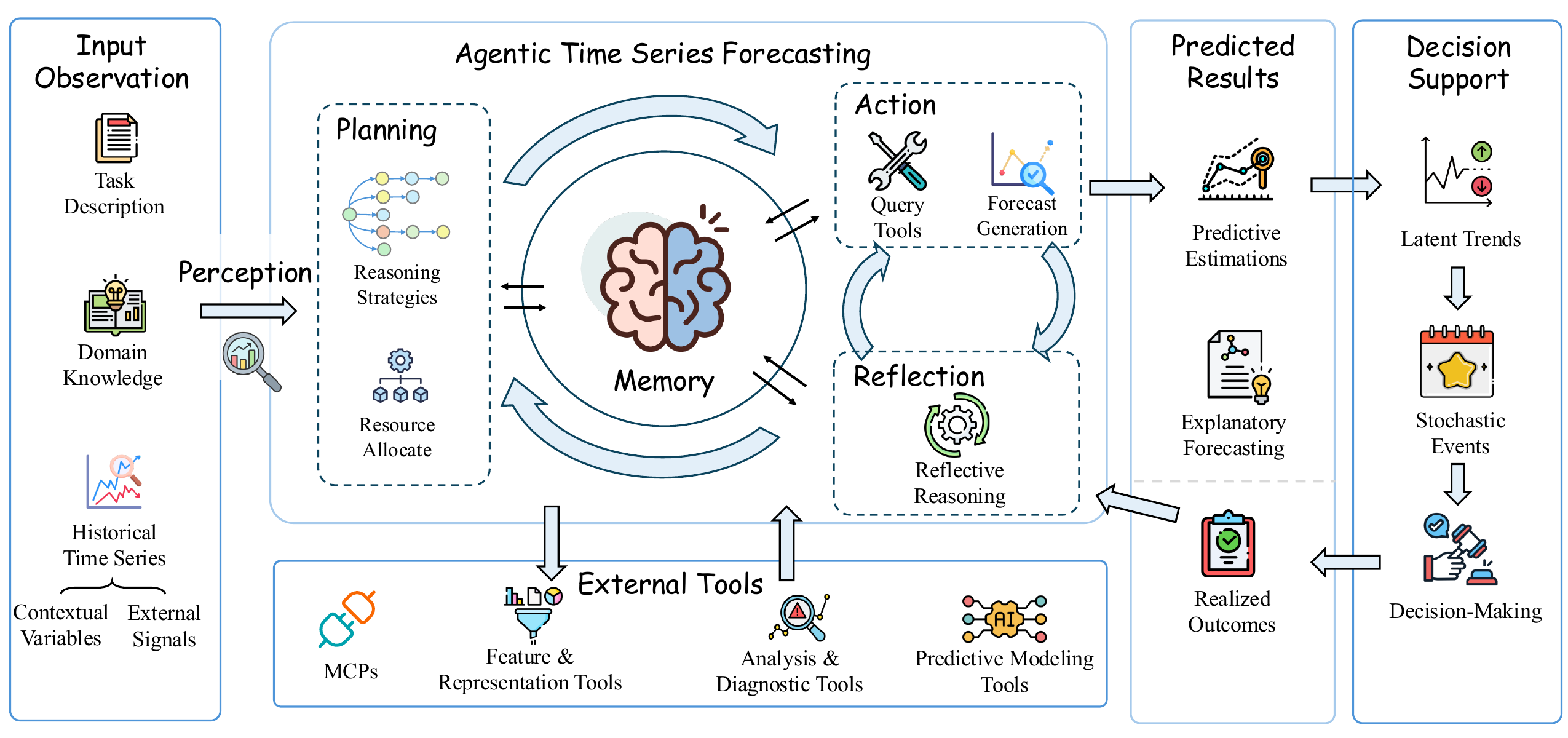} 
    \caption{Illustration of the core components and workflow of agentic time series forecasting.} 
    \label{fig:frameowork}
\end{figure*}
\subsection{From Model-Centric Forecasting to Agentic Forecasting}

Traditional time series forecasting is predominantly formulated under a model-centric paradigm, where forecasting is treated as a static, single-pass mapping from fixed historical observations to future values. Progress under this formulation is largely driven by predictive model design and optimization, while the forecasting process itself is assumed to be fully specified once inputs are given~\cite{liu2024time}. This perspective implicitly assumes that relevant information is available in advance, objectives are fixed, and predictions result from a single execution of a predictive function. ATSF departs from this view by reframing forecasting as a multi-turn, interactive decision-making process. Rather than treating forecasts as direct model outputs, ATSF views them as the result of a sequence of decisions involving perception, planning, action, reflection, and memory. Key aspects of forecasting—such as context selection, task decomposition, tool invocation, and prediction revision—are determined dynamically through cognitive reasoning and autonomous interaction~\cite{yao2022react}. This shift moves the focus of forecasting from model execution to process organization, where performance depends not only on the predictive capacity of individual models, but also on how forecasting activities are structured, coordinated, and adapted over time.

\subsection{Core Components of ATSF}
As illustrated in Figure~\ref{fig:frameowork}, ATSF views forecasting as an iterative decision-making process structured around perception, planning, action, reflection, and memory, rather than a single prediction function.

\paragraph{Perception.}
Perception refers to the capability of a forecasting system to extract task-relevant information from noisy, heterogeneous, and often unstructured inputs before any forecasting decision is made~\cite{wang2024survey}. In ATSF, perception is responsible for transforming raw observations into meaningful internal representations by filtering irrelevant signals, resolving distributional discrepancies, and identifying salient patterns that matter for the current forecasting objective. This process may involve data normalization across domains or instances~\cite{kim2021reversible}, as well as the preparation and selection of key features that support  reasoning.

\begin{positionbox}
\textbf{Our position:} \textit{Crucially, perception in ATSF is not merely a fixed preprocessing step, but a necessary and adaptive cognitive process that determines what information is perceived as relevant under different contexts and tasks. By explicitly modeling perception, ATSF acknowledges that forecasting performance depends not only on how predictions are generated, but also on how relevant information is perceived and structured from initially disorganized inputs.}
\end{positionbox}

\paragraph{Planning.}
Planning is a necessary precursor to action in ATSF, providing a global view of the forecasting task before any concrete execution takes place~\cite{huang2024understanding}. Given the perceived context, planning is responsible for formulating forecasting objectives, decomposing complex forecasting tasks into manageable sub-tasks, and determining high-level strategies that guide subsequent actions. Rather than directly producing predictions, planning defines what needs to be forecast, under which assumptions, and in what sequence, establishing a structure for the forecasting.
\begin{positionbox}
    \textbf{Our position:} \textit{A defining characteristic of planning in ATSF is its dynamic nature. Planning is not a one-time decision made prior to forecasting, but an ongoing process that can be revised in response to new observations, intermediate outcomes, or external feedback. Through dynamic replanning, agentic forecasting systems can adjust their goals, task decompositions, and strategies as conditions evolve, enabling flexible and responsive forecasting beyond fixed, static workflows.}
\end{positionbox}

\paragraph{Action.}
Action denotes the execution of decisions formulated during planning through autonomous interaction with tools~\cite{schick2023toolformer,cheng2026instructtime++}. In ATSF, actions are primarily responsible for preparing, enriching, and structuring the contextual evidence needed for effective forecasting. These actions may include invoking predictive models~\cite{cheng2025convtimenet}, performing statistical analyses, retrieving auxiliary information, or transforming intermediate representations. Notably, time series forecasting itself is treated as one type of action within a broader action space, rather than the sole objective.

\begin{figure}[t]  
    \centering
    \includegraphics[width=0.9\columnwidth]{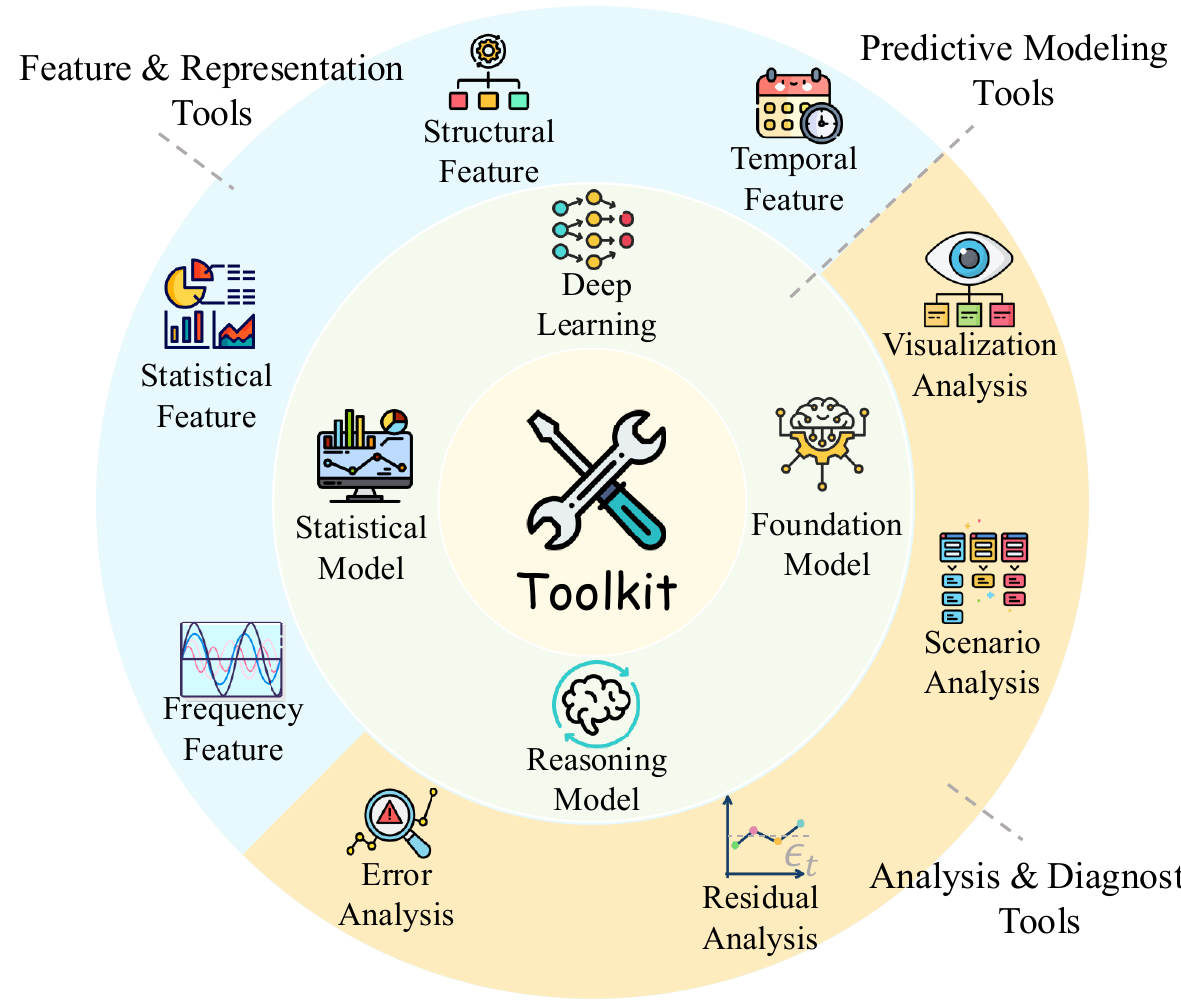} 
    \caption{Illustration of the representative used tool for agentic time series forecasting.} 
    \label{fig:tookit}
\end{figure}

By defining forecasting as an action, ATSF provides a natural mechanism for integrating diverse time series methods and tools. Since no single predictive model consistently performs best across all domains and scenarios~\cite{brigato2025position}, agentic actions allow forecasting systems to flexibly select, combine, or switch among traditional statistical models, machine learning approaches, and domain-specific techniques based on the current context. In this sense, action serves as the primary interface through which prior advances in time series forecasting are incorporated into an agentic framework, as exemplified by the representative toolkit in Figure~\ref{fig:tookit}.

\begin{positionbox}
    \textbf{Our position:} \textit{Actions in ATSF can be further characterized by two complementary execution patterns. Parallel actions execute multiple operations concurrently without intermediate reasoning, enabling efficient exploration or rapid context construction. In contrast, sequential actions interleave reasoning and execution, where intermediate outcomes inform subsequent decisions and actions. This distinction captures the spectrum from direct execution to deliberative, step-by-step interaction, reflecting different operational modes.}
\end{positionbox}

\paragraph{Reflection.} Reflection enables iterative forecasting correction by allowing an agentic forecasting system to evaluate, interpret, and revise its own predictions~\cite{gou2023critic}. Rather than serving as a terminal assessment step, reflection in ATSF functions as an internal self-evaluation mechanism that examines forecasting outcomes in light of expectations, assumptions, and contextual evidence. Through reflection, the system identifies discrepancies, assesses confidence and uncertainty, and determines whether corrective actions or alternative forecasting strategies are needed.
\begin{positionbox}
   \textbf{Our position:} \textit{A defining feature of reflection in ATSF is its support for self-reflection and self-judgment, whereby the system reasons about the quality and plausibility of its own forecasts without external supervision. This reflective process provides the basis for iterative prediction refinement, enabling forecasts to be adjusted, re-generated, or augmented through subsequent planning and action. By explicitly modeling reflection as a core component, ATSF moves beyond one-shot forecasting and supports a multi-turn forecasting process in which predictions can be progressively improved over time. }
\end{positionbox}

\paragraph{Memory.}
Memory enables ATSF to achieve cross-instance experience transfer~\cite{qiu2025alita,wu2025evolver}, continual evolution over time~\cite{cai2025flex,cao2025remember}. Unlike conventional approaches that rely solely on parameter updates to absorb past information, ATSF emphasizes explicit memory mechanisms that store reusable experience beyond individual forecasting instances. The role of memory is not to retain raw observations, but to preserve information with decision value—such as recurring patterns, effective strategies, failure cases, and contextual regularities—that can guide future forecasting decisions.

A defining property of memory in ATSF is its dynamic evolving nature. Memory is continuously updated, refined, and reorganized as new experiences are accumulated, allowing outdated or low-value information to be revised or discarded~\cite{tao2026memcast}. This dynamic evolution supports continual adaptation and prevents agentic forecasting systems from becoming overfitted to past conditions. Memory representations may take multiple forms, including textual descriptions, structured knowledge graphs~\cite{edge2024local}, or learned embeddings, reflecting different types of experiential knowledge.
\begin{positionbox}
    \textbf{Our position:} \textit{Memory in ATSF is typically accessed through retrieval mechanisms that allow relevant experience to be selectively recalled based on the current context, task, or uncertainty. Importantly, memory is inherently hierarchical, comprising multiple levels that capture experience at different granularities, such as instance-level outcomes, task-level strategies, and domain-level regularities. This layered structure enables agentic forecasting systems to generalize across samples and domains while retaining the flexibility to adapt to specific situations.}
\end{positionbox}

\subsection{ATSF as an Iterative Decision-Making Process}
ATSF formulates forecasting as an iterative decision-making process, in which predictions emerge from a sequence of interdependent decisions rather than a single computation. The objective of forecasting is thus not only to generate numerical outputs, but to continuously revise forecasting decisions in response to uncertainty, new information, and accumulated experience, fundamentally distinguishing ATSF from single-pass formulations. Each decision cycle in ATSF is structured around perception, planning, action, reflection, and memory. Perception organizes heterogeneous information into task-relevant representations. Planning defines forecasting objectives and strategies based on the perceived context. Action executes these decisions through autonomous tool use and forecasting operations, where prediction itself is treated as one action among many. Reflection evaluates outcomes to identify uncertainty and discrepancies, while memory accumulates experience across cycles to inform future decisions. Crucially, ATSF is iterative by design rather than post-hoc refinement. Reflection and memory feed back into planning, enabling dynamic replanning as assumptions change or new evidence arises. This iterative structure allows forecasting behavior to adapt over time, aligning agentic forecasting with expert reasoning and decision support in complex, non-stationary environments.

\section{Implementation Paradigms of ATSF}
Agentic forecasting admits multiple implementation paradigms, including workflow-based design, agentic reinforcement learning, and AgentFlow, which integrates structured workflows with localized reinforcement learning.	

\subsection{Agentic Workflow Design}
Workflow-based agentic forecasting, referred to here as the \emph{Workflow} paradigm, represents the most direct realization of agentic forecasting by explicitly structuring the forecasting process as a sequence of cognitive steps rather than a single predictive function~\cite{yao2022react,zhang2025alphacast}. In this paradigm, forecasting is organized around a predefined workflow that delineates planning, action, reflection, and memory as first-class stages, each responsible for a distinct decision role. By decomposing forecasting into interpretable and modular steps, workflow-based design enables adaptive context selection, tool invocation, and iterative revision without requiring end-to-end learning across the entire system. Importantly, this approach highlights that agentic behavior in forecasting does not primarily stem from model complexity, but from how forecasting activities are organized, coordinated, and evaluated within a structured decision process.
\begin{table}[t]
\centering
\small
\caption{Comparison of three agentic implementation strategies for time series forecasting. Hybrid methods aim to balance the stability of workflows with the adaptability of reinforcement learning.}
\label{tab:agentic_methods_comparison}
\resizebox{\columnwidth}{!}{
    \begin{tabular}{l >{\raggedright\arraybackslash}m{2.8cm} >{\raggedright\arraybackslash}m{2.6cm} >{\raggedright\arraybackslash}m{2.6cm}}
        \toprule
        \textbf{Method} & \textbf{Core Mechanism} & \textbf{Strengths} & \textbf{Weaknesses} \\
        \midrule
        \textbf{Workflow} & Structured execution via predefined DAGs or SOPs (e.g., LangGraph~\cite{wang2024agent}) & High interpretability, stability, and ease of debugging. & Limited flexibility for unseen scenarios; rigid process \\
        \midrule
        \textbf{AgenticRL} & Policy optimization via trial-and-error and reward feedback. & Autonomous evolution; discovers novel strategies & High training instability; sample inefficiency; hard to interpret \\
        \midrule
        \textbf{AgenticFlow} & Integration of explicit planning (Workflow) and implicit learning (RL/Memory). & Balances stability with adaptability; continuous improvement & High architectural complexity; difficult to tune \\
        \bottomrule
    \end{tabular}%
}
\end{table}
\subsection{Agentic Reinforcement Learning}
Agentic reinforcement learning, referred to here as the \emph{AgenticRL} paradigm, views forecasting as an interactive decision-making process in which an agent learns to improve forecasting behaviors through repeated interaction with its environment~\cite{shinn2023reflexion,jiang2025tablemind,cheng2025agent}. Rather than optimizing predictive models in isolation, this paradigm applies reinforcement learning to the decisions surrounding forecasting, such as how to plan forecasting objectives, select actions, and revise strategies based on feedback. By framing forecasting within a long-horizon optimization perspective, agentic reinforcement learning enables agents to account for delayed rewards, downstream consequences, and trade-offs between accuracy, robustness, and risk. Importantly, this does not imply replacing forecasting models with reinforcement learning, but rather using reinforcement learning to guide how forecasting decisions are made and revised. Crucially, this perspective emphasizes that learning in agentic forecasting targets the decision process itself, allowing forecasting systems to adapt over time as they accumulate experience from interaction.

\subsection{Hybrid Agentic Workflow}
Hybrid agentic workflow, referred to here as the \emph{AgenticFlow} paradigm, represents a unifying approach that integrates workflow-based design with agentic reinforcement learning through localized and structured adaptation~\cite{li2025flow}. In this paradigm, the overall forecasting workflow remains explicitly defined and interpretable, while learning is selectively applied to specific decision points within the workflow, such as context selection, strategy switching, or revision triggering. By constraining reinforcement learning to local adaptations rather than end-to-end optimization, AgentFlow balances flexibility with stability, enabling forecasting systems to evolve without sacrificing reliability or transparency. This integration highlights that effective agentic forecasting emerges not from unconstrained learning, but from the principled coordination of structured processes and adaptive decision-making.

\section{Alternative Views and Discussion}
ATSF can be viewed as a conceptual reframing of forecasting rather than a replacement for existing modeling approaches. This section situates ATSF among related perspectives and clarifies its scope and applicability. One related view is that recent advances in large-scale and foundation models already address many challenges in forecasting. While such models improve predictive capacity through scale and representation learning, they largely operate in a single-pass execution paradigm. ATSF is orthogonal to model scale: it reframes forecasting as an explicit decision-making process, in which planning, reflection, and memory shape forecasting behavior over time without requiring changes to underlying predictive models. Another perspective interprets agentic forecasting as an instance of complex forecasting pipelines or automated workflows. Conventional pipelines follow static execution graphs with predefined operations. In contrast, ATSF treats forecasting as a dynamic decision process, where context selection, tool use, and revision are adaptively controlled through iterative planning and reflection. ATSF is also related to online learning and adaptive modeling, which address non-stationarity through incremental parameter updates. While complementary, these methods focus on adapting models rather than forecasting processes. By introducing explicit memory and reflection, ATSF enables experience accumulation, strategy revision, and behavioral adaptation beyond parameter-level updates.

	\section{Opportunities and Challenges}
	The opportunities and challenges discussed below jointly characterize the promise and complexity of agentic forecasting, highlighting both its transformative potential and the obstacles that must be addressed to realize it in practice.
	
\subsection{Opportunities}
Viewing time series forecasting through an agentic lens reveals a range of research opportunities that extend beyond improving predictive accuracy, reshaping how forecasting systems are designed, learned, and deployed.
\paragraph{Opportunity 1: From Model Iteration to System and Tool Evolution.}
   Agentic forecasting shifts the primary axis of progress in time series forecasting from model-centric iteration to system- and tool-level evolution. Instead of focusing solely on architectures, agentic frameworks emphasize how forecasting components are organized, orchestrated, and evaluated within a broader decision process. This shift enables progress through workflow design, tool composition, and decision policies, allowing forecasting systems to improve even when underlying models remain unchanged. Consequently, innovation extends beyond architectural novelty to principled system-level design.	

   \paragraph{Opportunity 2: Integrating Heterogeneous Learning Paradigms.}
By treating forecasting as an agentic process, ATSF provides a natural interface for integrating heterogeneous learning paradigms, including large models, small predictive models, and domain-specific tools. Large models can support planning, reasoning, and reflection, while smaller models offer efficiency, stability, and reliable numerical prediction. Agentic forecasting enables these components to collaborate rather than compete, allowing different models to contribute complementary capabilities within a unified workflow. This integration opens new avenues for leveraging prior advances in time series modeling without requiring their replacement.

\paragraph{Opportunity 3: Modeling Expert-Like Forecasting Behavior.}
Agentic forecasting offers an opportunity to explicitly model forecasting behaviors that are commonly exhibited by human experts but largely absent from traditional forecasting systems. In practice, expert forecasting is iterative, hypothesis-driven, and grounded in experience, involving repeated examination of historical patterns, contextual signals, and prior outcomes. By embedding planning, reflection, and memory into the forecasting process, ATSF makes such expert-like behaviors explicit, systematic, and reproducible. This alignment with human cognitive practices provides a principled path toward more interpretable and trustworthy forecasting systems.

\paragraph{Opportunity 4: Forecasting under Complex and Dynamic Scenarios.}
Many real-world forecasting scenarios are characterized by non-stationarity, incomplete observations, and event-driven disruptions, where single-pass prediction often proves insufficient. Agentic forecasting is inherently well-suited to these settings, as its iterative and interactive nature allows forecasts to be revised in response to new information and changing conditions. By enabling dynamic context selection, adaptive strategy adjustment, and continual reassessment, ATSF opens new research directions for forecasting in open-world and high-stakes environments that challenge traditional model-centric approaches.

\paragraph{Opportunity 5: Human–Agent Collaborative Decision-Making.}
Reframing forecasting as an agentic process creates new opportunities for human–agent collaboration in decision-making. Rather than treating human intervention as an external correction mechanism, agentic forecasting incorporates human input as part of the forecasting loop, allowing preferences, constraints, and domain knowledge to inform planning and evaluation. This interactive paradigm supports collaborative forecasting workflows in which human expertise and algorithmic reasoning jointly shape outcomes. Such collaboration is particularly valuable in high-impact domains, where accountability, interpretability, and shared responsibility are essential.

\subsection{Challenges}
These opportunities, however, do not come for free. Realizing the promise of agentic forecasting introduces a set of fundamental challenges that span learning, system design, and deployment.
    
\paragraph{Challenge 1: Memory Design for Experience Accumulation and Transfer.}
Memory is a central yet underexplored component of agentic forecasting. Unlike conventional forecasting models that implicitly encode past information in parameters, agentic systems require explicit mechanisms to represent, store, and retrieve forecasting experience. Key challenges include determining what constitutes useful experience, how to generalize across contexts without amplifying spurious patterns, and how to prevent outdated or erroneous memories from degrading future decisions. Designing memory structures that support accumulation, abstraction, and transfer of forecasting knowledge remains a fundamental research problem.

\paragraph{Challenge 2: Toolkit Infrastructure and Standardization.}
Agentic forecasting relies on the coordinated use of heterogeneous tools, including predictive models, analytical procedures, and external information sources. However, the lack of standardized interfaces, semantics, and verification mechanisms poses a significant challenge. Without principled tool abstractions, agentic systems risk becoming brittle, opaque, or difficult to extend. Establishing modular, composable, and verifiable toolkits is essential for enabling scalable and reliable agentic forecasting systems.

\paragraph{Challenge 3: Multi-Agent Coordination.}
As forecasting tasks grow in complexity, multiple agents may be required to operate across different horizons, objectives, or sources of uncertainty. Coordinating such agents introduces challenges related to role allocation, information sharing, and conflict resolution. Moreover, assigning credit or responsibility among agents when forecasting outcomes improve or deteriorate remains non-trivial. Addressing these coordination and credit assignment issues is critical for realizing the potential of multi-agent forecasting systems.

\paragraph{Challenge 4: Reliability, Uncertainty, and Stability.}
Ensuring reliable behavior under uncertainty is a core challenge for agentic forecasting. Iterative reasoning and multi-step interaction can amplify uncertainty, propagate errors, or lead to unstable decision trajectories. While uncertainty estimation is well studied at the model level, its role in guiding exploration, reflection, and revision within agentic workflows is less understood. Developing principled mechanisms for uncertainty-aware reasoning is essential for deploying agentic forecasting in high-stakes settings.

\paragraph{Challenge 5: Efficiency and Scalability.}
Agentic forecasting introduces additional computational overhead due to iterative reasoning, interaction, and coordination. Balancing the benefits of agentic behavior with efficiency and scalability constraints is therefore a critical challenge. Not all forecasting decisions require the same level of reasoning or adaptation, suggesting the need for selective and resource-aware agentic mechanisms. Designing systems that allocate computational effort judiciously while preserving forecasting quality remains an open research question.

\paragraph{Challenge 6: Safety, Privacy, and Deployment Constraints.}
Deploying agentic forecasting systems in real-world environments raises important safety and privacy concerns. In domains such as healthcare, energy, and finance, forecasting systems must operate under strict regulatory and data protection constraints. Ensuring that agentic behaviors remain auditable, controllable, and compatible with privacy-preserving deployment is a non-trivial challenge. These considerations extend beyond algorithm design to encompass system architecture and operational protocols.

\paragraph{Challenge 7: From Prediction to Decision Accountability.}
Agentic forecasting blurs the boundary between prediction and decision-making, raising questions of accountability and responsibility. As forecasts increasingly influence or automate downstream actions, it becomes unclear how errors should be attributed among models, agents, and human stakeholders. Defining clear principles for responsibility allocation and decision accountability is essential for trustworthy deployment. Addressing this challenge requires rethinking evaluation criteria and governance frameworks alongside technical advances.

\section{Conclusion}

This position paper highlights the potential of ATSF to advance research and practice beyond model-centric prediction. We argue that forecasting should be viewed not as a static, single-pass estimation task, but as an agentic process grounded in planning, action, reflection, and memory, where performance emerges through iterative decision-making and adaptation. By articulating workflow-based design, agentic reinforcement learning, and the AgentFlow paradigm, we contend that organizing forecasting as an agentic process enables more flexible, interpretable, and decision-aware systems. Despite ongoing challenges in memory design, reliability, efficiency, and deployment, agentic forecasting represents a meaningful step toward aligning forecasting systems with real-world decision processes. Looking ahead, we encourage the community to develop principled agentic workflows, adaptive forecasting mechanisms, and evaluation frameworks that capture the iterative nature of forecasting.

    
	\newpage

	\bibliography{icml26}
	\bibliographystyle{icml2026}
	
	\newpage
	\appendix
	\onecolumn
	
	
	
\end{document}